\documentclass[letterpaper, 10 pt, conference]{ieeeconf}  

\IEEEoverridecommandlockouts                              

\usepackage{comment}
\usepackage{gensymb}
\usepackage{array}
\usepackage{multicol}
\usepackage{multirow}
\usepackage{graphicx}
\usepackage{svg}
\usepackage{amsmath, amsfonts}
\usepackage{bm}
\usepackage{booktabs}

\DeclareMathOperator*{\argmax}{arg\,max}

\newcommand{\RNum}[1]{\uppercase\expandafter{\romannumeral #1\relax}}
\title{\LARGE \bf
Sound Source Localization for\\Human-Robot Interaction in Outdoor Environments
}

\author{
    Victor Liu$^*$
    \and Timothy Du$^*$\thanks{Victor Liu and Timothy Du are both lead authors and contributed equally to the work}
    \and Jordy Sehn$^{++}$\thanks{$^{++}$Jordy Sehn works for Amtech Aeronautical Ltd.}
\and Jack Collier$^*$\thanks{$^*$Defence Research and Development Canada - Suffield Research Centre}
    \and Fran\c{c}ois Grondin$^+$\thanks{$^+$Department of Electrical Engineering and Computer Engineering, Interdisciplinary Institute for Technological Innovation (3IT), 3000 boul. de l'Universit\'e, Universit\'e de Sherbrooke, Sherbrooke, Qu\'ebec (Canada) J1K 0A5. Corresponding author: francois.grondin2@usherbrooke.ca.}
}

\begin{document}

\maketitle
\thispagestyle{empty}
\pagestyle{empty}

\begin{abstract}

This paper presents a sound source localization strategy that relies on a microphone array embedded in an unmanned ground vehicle and an asynchronous close-talking microphone near the operator.
A signal coarse alignment strategy is combined with a time-domain acoustic echo cancellation algorithm to estimate a time-frequency ideal ratio mask to isolate the target speech from interferences and environmental noise.
This allows selective sound source localization, and provides the robot with the direction of arrival of sound from the active operator, which enables rich interaction in noisy scenarios.
Results demonstrate an average angle error of 4 degrees and an accuracy within 5 degrees of 95\% at a signal-to-noise ratio of 1dB, which is significantly superior to the state-of-the-art localization methods.

\end{abstract}

\section{INTRODUCTION}

For robotics operations, voice control is an effective technology for human-robot interaction \cite{yen2024drone,nakadai2024swarm,grondin2022odas}. It is intuitive, hands-free, and offers a low cognitive burden to the operator. This is important in high-stress scenarios, such as defense and security applications, where the operator needs to interact with the robot while maintaining environmental situational awareness and communicating, via headset or gesture, with other personnel. Advances in natural language processing (NLP) with large language models (LLM) have accelerated the adoption of voice control technologies \cite{fathullah2024prompting}, as these models are able to accurately interpret commands from diverse set of natural language statements. 

For many outdoor field robotics applications, an essential problem with voice control is filtering ambient noise and voices from the true operator voice commands \cite{lagace2023ego}. Sound Source Localization (SSL) is one strategy for dealing with this issue whereby the direction of commands can be determined and used to verify the end user with another sensor such as lidar \cite{chamorro2021neural}. Additionally, the authors envision that output from an SSL algorithm can provide additional context for a robot command. For instance, an operator issuing the command "follow me" can use the SSL output to determine which where to turn to begin following the operator. Directional output can also allow for voice commands to be combined with gesture signals, detected from optical or lidar sensors, to allow for complex commands like "Go over there" to be understood. Finally, in a multi-user control situation, the SSL output can be used to verify that the command is coming from the operator and can assist in operator handover.

SSL can be achieved using Multiple Signal Classification (MUSIC) and Steered Response Power Phase Transform (SRP-PHAT) methods.
MUSIC is based on Standard Eigenvalue Decomposition (SEVD-MUSIC), formerly used for narrowband signals \cite{schmidt1986multiple}, and adapted to broadband speech signals \cite{ishi2009evaluation}.
It however assumes that speech is more powerful than interference, which is usually not the case.
Variants such as the Generalized Eigenvalue Decomposition (GEVD-MUSIC) \cite{nakamura2009intelligent} and the Generalized Singular Value Decomposition (GSVD-MUSIC) \cite{nakamura2012real} methods tackle this problem.
However, all MUSIC-based methods rely on online eigenvalue or singular value decompositions, which are computationally expensive.
SRP-PHAT (implemented using Generalized Cross-Correlation (GCC-PHAT)) uses fewer computations than MUSIC \cite{grondin2019lightweight} but is sensitive to background noise.
Difference Singular Value Decomposition with Phase Transform (DSVD-PHAT) is proposed as an alternative to SRP-PHAT to improve noise robustness, and also reduce the algorithm complexity when searching for a direction of arrival (DoA) in 3D \cite{grondin2019fast}.

In this work, we propose to use the Minimum Variance Distortionless Response (MVDR) beamformer to localize a target signal and ignore interfering sources.
This approach is inspired by the improved MVDR that relies on spatial covariance matrices estimated with masks \cite{erdogan2016improved}.
For this application, the mask are estimated with acoustic echo cancellation \cite{kuech2014state, paleologu2013study} based on a clean reference signal provided by a close-talking microphone on the operator. The use of a close-talking microphone is motivated by military scenarios where personnel will already have a communications microphone that can be exploited.
We demonstrate that this strategy makes the system robust to noise, provides accurate SSL results, and can run in real-time on an onboard computer.

\section{SOUND SOURCE LOCALIZATION SYSTEM}

\begin{figure}[!ht]
    \centering
    \vspace{6pt}
    \includegraphics[width=\linewidth]{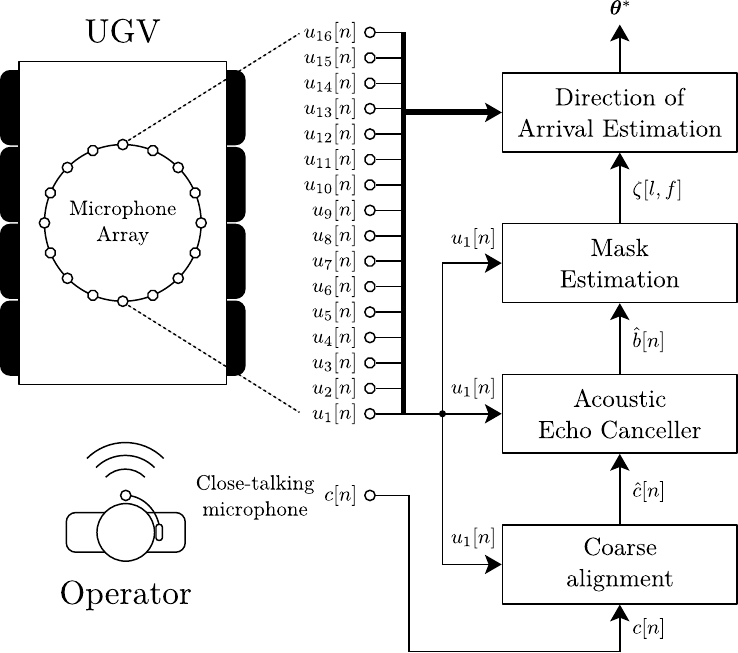}
    \caption{Overview of the proposed method. The UGV is equipped with a 16-microphone array, and the operator uses a close-talking microphone.}
    \label{fig:sysoverview}
\end{figure}

Figure \ref{fig:sysoverview} shows the proposed system that consists of an unmanned ground vehicle (UGV) equipped with a cirular 16-microphone array (MA).
The operator uses a close-talking microphone with a push-to-talk button to give speech commands to the UGV.
A wireless transmitter sends this speech signal to the UGV.
The coarse alignment module realigns it to cope with transmission delays.
Using one of the microphone array signals, the noise signal (i.e. the mixture minus the speech signal) is estimated using an acoustic echo canceller.
This provides a time-frequency mask, that can be used for selective direction of arrival estimation.

\subsection{Signal model}

The expression $s[n] \in \mathbb{R}$ stands for the target sound source in the discrete time domain ($n \in \mathbb{Z})$ for a given sample rate, $f_S \in \mathbb{N}$, in samples/sec.
The sound is captured by each microphone $m \in \{1, 2, \dots, M\}$ on the UGV (with $M \in \mathbb{N}$ being the number of microphones), where $h_{ugv,m}[n] \in \mathbb{R}$ stands for the impulse response that models the propagation of sound from source to microphone, and $b_m[n] \in \mathbb{R}$ is the background noise.
The sound captured at each microphone on the robot is given by the following:
\begin{equation}
    u_m[n] = h_{ugv,m}[n] * s[n] + b_m[n],
\end{equation}
where $*$ stands for the linear convolution operator.
A close-talking microphone also captures the voice of the operator (denoted as $c[n] \in \mathbb{R}$). The corresponding signal model includes the microphone transfer function ($h_{close}[n] \in \mathbb{R}$) and ignores background noise:
\begin{equation}
    c[n] = h_{close}[n] * s[n].
\end{equation}

\subsection{Coarse alignment}

The close-talking signal $c[n]$ is captured asynchronously with respect to the microphone array signals $u_m[n]\ \forall\ m$, and transmitted using a wireless connection to the UGV.
The first step involves coarse alignment to eliminate the random delay introduced by the wireless communication protocol.
The Short-Time Fourier Transform (STFT) is computed with a Hann window of size  $N \in \mathbb{N}$ and a hop length of $\Delta N \in \mathbb{N}$ samples, for all signals as follows: 
\begin{equation}
U_m[l,f] = \mathrm{STFT}\{u_m[n]\},
\label{eq:stft_um}
\end{equation}
\begin{equation}
C[l,f] = \mathrm{STFT}\{c[n]\},
\label{eq:stft_c}
\end{equation}
where $l \in \{0, 1, \dots, L-1\}$ stands for the frame index (for a total of $L \in \mathbb{N}$ frames), and $f \in \{0, 1, \dots, F-1\}$ represents the frequency bin index (for a total of $F \in \mathbb{N}$ bins).
The cross-correlation between both spectrograms is then obtained with respect to the frame index using a frequency-wise generalized cross-correlation with phase-transform (GCC-PHAT) as follows:
\begin{equation}
    r_{UC}[\tau,f] = \sum_{k=0}^{L-1}{\frac{R_C[k,f]R_U[k,f]^*}{|R_C[k,f]||R_U[k,f]|}e^{j2\pi \tau k/L}},
\end{equation}
with $\{\dots\}^*$ that denotes the complex conjugate operator, $|\dots|$ that extracts the magnitude, and where:
\begin{equation}
    R_U[k,f] = \sum_{l=0}^{L-1}{|U_1[l,f]|^{\beta}}e^{-j2\pi kl/L},
\end{equation}
\begin{equation}
    R_C[k,f] = \sum_{l=0}^{L-1}{|C[l,f]|^{\beta}}e^{-j2\pi kl/L},
\end{equation}
with $\beta \in [0,1]$ to compress the amplitude in with a logarithmic scale for high amplitudes and a linear scale for small values.
The optimal frame delay $\tau^*$ corresponds to the value of $\tau$ that maximizes the sum of cross-correlations accross all frequency bins:
\begin{equation}
    \tau^* = \argmax_{\tau}\left\{\sum_{f=0}^{F-1}{r_{UC}[\tau,f]}\right\}
\end{equation}

This leads to the coarse-aligned close-talking microphone signal in the frequency domain:
\begin{equation}
    \hat{C}[l,f] = C[l+\tau^*,f],
\end{equation}
which is then converted back to the time-domain using the inverse STFT:
\begin{equation}
    \hat{c}[n] = \mathrm{STFT}^{-1}\{\hat{C}[l,f]\}.
\end{equation}

\subsection{Acoustic echo cancellation}

An acoustic echo canceller (AEC) based on the Kalman filter in the time-domain performs fine alignment of the close-talking signal and remove it from the reference signal on the UGV, to estimate the background noise.
This method, inspired from a demo code available online\footnote{https://github.com/ewan-xu/pyaec}, estimates $D \in \mathbb{N}$ coefficients of a finite impulse response (FIR) filter, denoted as the \emph{a priori} state mean ($\hat{\mathbf{x}}_{n|n-1} \in \mathbb{R}^{D \times 1}$) and the \emph{a posteriori} ($\hat{\mathbf{x}}_{n|n} \in \mathbb{R}^{D \times 1}$) state mean.
The state transition, or prediction, obeys this linear equation:
\begin{equation}
    \tilde{\mathbf{x}}_{n|n-1} = \mathbf{F}_n \tilde{\mathbf{x}}_{n-1|n-1},
\end{equation}
where the transition model $\mathbf{F}_n \in \mathbb{R}^{D \times D}$ is time independent and assumes stationnary states, such that $\mathbf{F}_n = \mathbf{I}_D$, with $\mathbf{I}_D$ that denotes the identity matrix of size $D$.
As the process noise is assumed to be Gaussian, the \emph{a priori} and \emph{a posteriori} state covariances are represented by $\mathbf{P}_{n|n-1}, \mathbf{P}_{n|n} \in \mathbb{R}^{D \times D}$, respectively.
The \emph{a priori} state covariance gets predicted as follows:
\begin{equation}
    \mathbf{P}_{n|n-1} = \mathbf{F}_n \mathbf{P}_{n-1|n-1} \mathbf{F}_n^T + \mathbf{Q}_n,
\end{equation}
where the process noise covariance matrix $\mathbf{Q}_n \in \mathbb{R}^{D \times D}$ is also time independent, and modeled as $\mathbf{Q}_n = \sigma^2 \mathbf{I}_D$.

State mean and covariance are updated at each sample in the time domain, as the goal is to minimize the measurement pre- and post-fit residuals, given by the expressions $\tilde{\mathbf{y}}_{n}, \tilde{\mathbf{y}}_{n|n} \in \mathbb{R}^{1 \times 1}$, respectively.
The pre-fit residual is given by:
\begin{equation}
    \tilde{\mathbf{y}}_n = \mathbf{z}_n - \mathbf{H}_n \hat{\mathbf{x}}_{n|n-1},
\end{equation}
where $\mathbf{z}_n \in \mathbb{R}^{1 \times 1}$ is the measurement, given by one of the microphone array signal on the UGV, here denoted by $u_1[n]$ if the first microphone is used. 
To perform FIR filtering with the estimated coefficients, the time dependent observation model $\mathbf{H}_n \in \mathbb{R}^{1 \times D}$ consists of the current sample and the $D-1$ last samples from the coarse-aligned close-talking microphone:
\begin{equation}
    \mathbf{H}_n = \left[
    \begin{array}{cccc}
    \hat{x}[n] & \hat{x}[n-1] & \dots & \hat{x}[n-D+1]
    \end{array}
    \right].
\end{equation}

Similarly, the pre-fit residual covariance $\mathbf{S}_n \in \mathbb{R}^{1 \times 1}$ corresponds to:
\begin{equation}
    \mathbf{S}_n = \mathbf{H}_n \mathbf{P}_{n|n-1} \mathbf{H}_n^T + \mathbf{R}_n,
\end{equation}
where the observation noise has the covariance $\mathbf{R}_n \in \mathbb{R}^{1 \times 1}$, where $\mathbf{R}_n = \|\tilde{\mathbf{y}}_n\|^2_2$.
The updated \emph{a posteriori} state estimate ($\tilde{\mathbf{x}}_{n|n}$) and estimate covariances ($\mathbf{P}_{n|n}$) are given by:
\begin{equation}
    \tilde{\mathbf{x}}_{n|n} = \tilde{\mathbf{x}}_{n|n-1} + \mathbf{K}_n \tilde{\mathbf{y}}_n,
\end{equation}
\begin{equation}
    \mathbf{P}_{n|n} = (\mathbf{I}_D - \mathbf{K}_n \mathbf{H}_n) \mathbf{P}_{n|n-1},
\end{equation}
where the optimal Kalman gain $\mathbf{K}_{n} \in \mathbb{R}^{D \times 1}$ is:
\begin{equation}
    \mathbf{K}_n = \mathbf{P}_{n|n-1} \mathbf{H}_n^T \mathbf{S}_n^{-1}.
\end{equation}

    Finally, the post-fit residual $\tilde{\mathbf{y}}_{n|n} \in \mathbb{R}^{1 \times 1}$ corresponds to:
\begin{equation}
    \tilde{\mathbf{y}}_{n|n} = \mathbf{z}_n - \mathbf{H}_n \hat{\mathbf{x}}_{n|n},
\end{equation}
and is used as the background noise estimate at each sample $n$, denoted as $\hat{b}[n] \in \mathbb{R}$.

\subsection{Mask estimation}

The estimated noise $\hat{b}[n]$ is transformed to the time-frequency domain using a STFT with the same parameters as in (\ref{eq:stft_um}) and (\ref{eq:stft_c}):
\begin{equation}
    \hat{B}[l,f] = \mathrm{STFT}\{\hat{b}[n]\}.
\end{equation}

The ideal time-frequency ratio mask (IRM) $\zeta[l,f] \in [0,1]^{L \times F}$ then emphasizes on bins dominated by speech:
\begin{equation}
    \zeta[l,f] = \frac{|\hat{X}[l,f]|^2}{|\hat{X}[l,f]|^2 + |\hat{B}[l,f]|^2}.
    \label{eq:zeta}
\end{equation}

It was found that the mask tends to overestimate the number of relevant bins to be used for localization, and this can degrade performances.
To deal with this, a modified mask is proposed:
\begin{equation}
    \hat{\zeta}[l,f] = \frac{|\hat{X}[l,f]|^2}{|\hat{X}[l,f]|^2 + |\hat{B}[l,f]|^2 + \displaystyle\frac{\alpha}{L}\sum_{k=0}^{L-1}{|\hat{B}[k,f]|^2}},
    \label{eq:zetahat}
\end{equation}
which ensures a minimum noise floor.

\subsection{Direction of arrival estimation}

The direction of arrival (DoA) estimation relies on the spatial covariance matrices (SCMs) for target speech and background noise, $\Phi_{XX}[f] \in \mathbb{C}^{M \times M}$ and $\Phi_{BB}[f] \in \mathbb{C}^{M \times M}$, computed for each frequency bin $f$:
\begin{equation}
    \Phi_{SS}[f] = \sum_{l=0}^{L-1}{\zeta[l,f]\mathbf{U}[l,f]\mathbf{U}[l,f]^H},
\end{equation}
\begin{equation}
    \Phi_{BB}[f] = \sum_{l=0}^{L-1}{(1-\zeta[l,f])\mathbf{U}[l,f]\mathbf{U}[l,f]^H},
\end{equation}
where the time-frequency signals are concatenated for all microphones in the vector $\mathbf{U}[l,f] \in \mathbb{C}^{M \times 1}$:
\begin{equation}
    \mathbf{U}[l,f] = \left[ 
    \begin{array}{cccc}
    U_1[l,f] & U_2[l,f] & \dots & U_M[l,f] \\
    \end{array}
    \right]^T.
\end{equation}

A new matrix is obtained as follows (which is similar to what is done with Minimum-Variance Distortionless Response (MVDR) beamforming):
\begin{equation}
    \Phi_{BB}^{-1}[f]\Phi_{SS}[f] = \left[
    \begin{array}{ccc}
    \phi_{1,1}[f] & \dots & \phi_{1,M}[f] \\
    \vdots & \ddots & \vdots \\
    \phi_{M,1}[f] & \dots & \phi_{M,M}[f] \\
    \end{array}
    \right].
\end{equation}

The noise SCM can be diagonally loaded if needed to avoid singularity.
Each potential direction of arrival of sound is scanned using a steered-response power beamformer with phase transform (SRP-PHAT) for each pair of microphones:
\begin{equation}
    E[\bm{\theta}] = \sum_{p=1}^{M}\sum_{q=p+1}^{M}{\Re\left\{\sum_{f=0}^{F-1}{\frac{\phi_{p,q}[f]}{|\phi_{p,q}[f]|}W_{p,q}[\bm{\theta},f]}\right\}},
\end{equation}
where $\Re\{\dots\}$ extracts the real part. The steering vector assumes free field propagation:
\begin{equation}
    W_{p,q}[\bm{\theta},f] = \exp\left\{\displaystyle\left(\frac{f_S}{c}\right)\left(\frac{2\pi f}{N}\right)(\mathbf{d}_p-\mathbf{d}_q)\cdot\bm{\theta}\right\},
\end{equation}
where $\bm{\theta} \in \{\mathbf{v} \in \mathbb{R}^3\ |\ \|\mathbf{v}\| = 1\}$ points in the direction of the source, $c$ stands for the speed of sound (in m/sec), and $\mathbf{d}_p \in \mathbb{R}^3$ and $\mathbf{d}_q \in \mathbb{R}^3$ stand for the positions (in m) of microphones $p$ and $q$.
The DoA corresponds to the direction that maximizes the power:
\begin{equation}
    \bm{\theta}^* = \argmax_{\bm{\theta}}{\{E[\bm{\theta}]\}}.
\end{equation}

\section{EXPERIMENTAL SETUP}

\begin{figure}[thpb]
      \begin{center}
      \includegraphics[width=\linewidth]{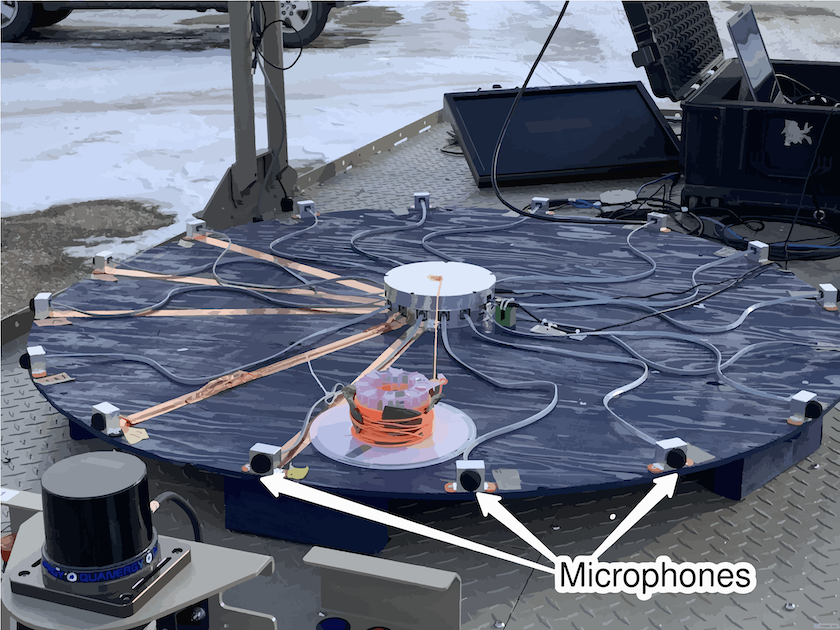}
      \caption{Argo robot microphone array used for trials. The distance from each microphone to the center of the array is 51.6 cm or 20.3 inches.}
      \label{fig::argomicarray}
      \end{center}
\end{figure}

\begin{figure}[thpb]
      \begin{center}
      \vspace{6pt}
      \includegraphics[width=\linewidth]{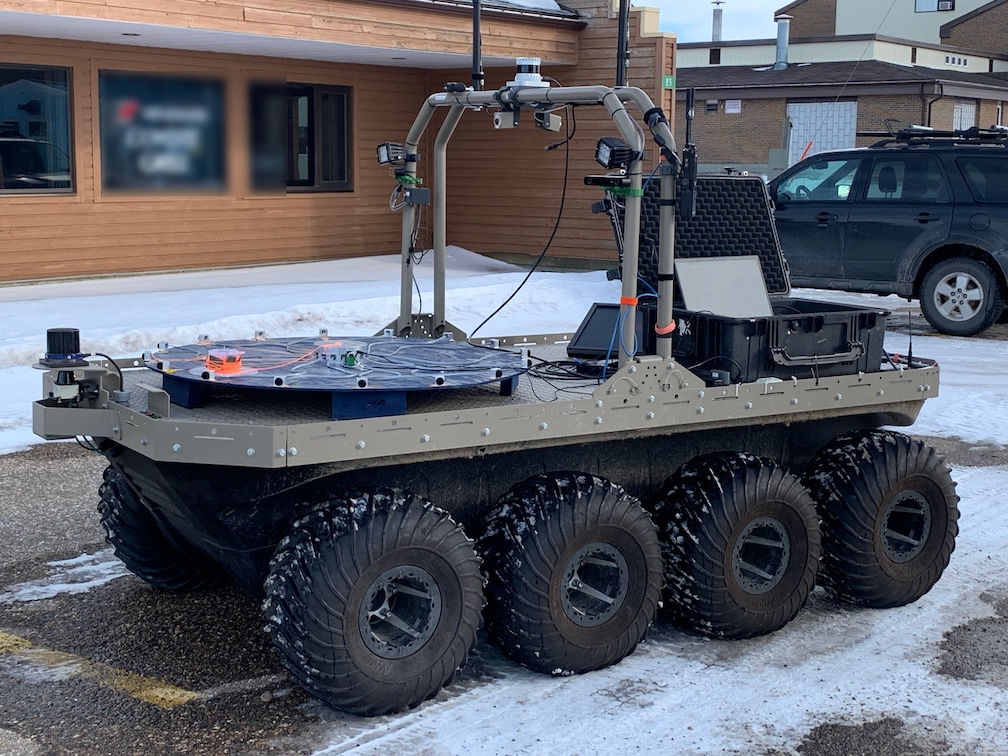}
      \caption{Position of Argo UGV robot during experimental data collection. The laptop computer on the back of the robot is connected to the Microphone array to capture the noisy audio.}
      \label{fig::argoposition}
      \end{center}
\end{figure}

Figures \ref{fig::argomicarray} and {\ref{fig::argoposition}} show the 16 channel MA and Argo J8 robot used in the experiments. The 16 omnidirectional microphones are located at the front of the robot at a consistent height. The MA is connected to the on board computer via USB, which runs the pipeline written in ROS and Python. Table \ref{tab:parameters} lists the parameters used for these experiments.

\begin{table}[!ht]
    \centering
    \vspace{4pt}
    \caption{Parameters used in experiments}
    \begin{tabular}{ccc}
    \toprule
        Parameter & Description & Value \\
    \midrule
        $f_S$ & Sample rate (samples/sec) & $16000$ \\
        $c$ & Speed of sound (m/sec) & $343$ \\
        $N$ & Frame size (samples) & $512$ \\
        $\Delta N$ & Hop size (samples) & $128$ \\
        $\alpha$ & Noise floor level & $0.05$ \\
        $\beta$ & Amplitude compression factor & $0.25$ \\
        $\sigma$ & Kalman filter process noise & $10^{-4}$ \\
    \bottomrule
    \end{tabular}
    \label{tab:parameters}
\end{table}

Prior to the experiment, a male and female recording of the phrase ``Make me the leader'' were recorded by the close-talking microphone attached to the on board Argo computer; these audio files served as the voice of the operator. Experiments were conducted in an outdoor parking lot next to the lab, with background noise produced by nearby buildings to achieve a realistic setup. The Argo was stationary for the duration of all trials. Trials were conducted at 8 distances between 3m and 10m, with increments of 1m. At each distance, 16 angles beginning at 0$\degree$ and increasing by 22.5$\degree$ at each iteration were tested. At each angle, every operator voice was broadcasted 3 times by a loud speaker with a sound level set at 10dB, for a total of 6 trials. This cumulated to a total of 96 experiments at each distance and 768 in total. 

For evaluation purposes, a separate group of 10 people were recorded by the microphone array and the close talking microphone saying the same phrase ``Make me the leader". The microphone array recordings were done at a distance of 3m and at 16 evenly distributed angles. In addition, the same group of people were recorded saying 10 different phrases identified as non-target speech (NTS) 1 through 10 by the microphone array at a distance of 3m as well. The 10 speakers were chosen from various demographic backgrounds and age to test pipeline robustness. Finally, a sample 16 channel recording of white noise (WN) is for the interference tests that are discussed in Section \ref{sec:results}.

\section{RESULTS AND DISCUSSION}
\label{sec:results}

The metric used to analyze the performance of the proposed pipeline is average model error ($\degree$) from the operator and model accuracy ($\%$) within 5$\degree$ of the operator. These metrics are evaluated under two series of tests; static and interference. Static tests consist of an operator recording from the close microphone along with the corresponding recording from the microphone array. Interference tests are comprised of an operator recording from the close microphone and an audio mixture created by mixing the corresponding microphone array recording with an arbitrary source of non-target speech (NTS) and white noise (WN). The mixtures are amplified by 10dB which showed on average a similar sound level relative to the microphone array recordings.

\subsection{Static}

\begin{figure}[thpb]
      \begin{center}
      \vspace{6pt}
      \includegraphics[width=\linewidth]{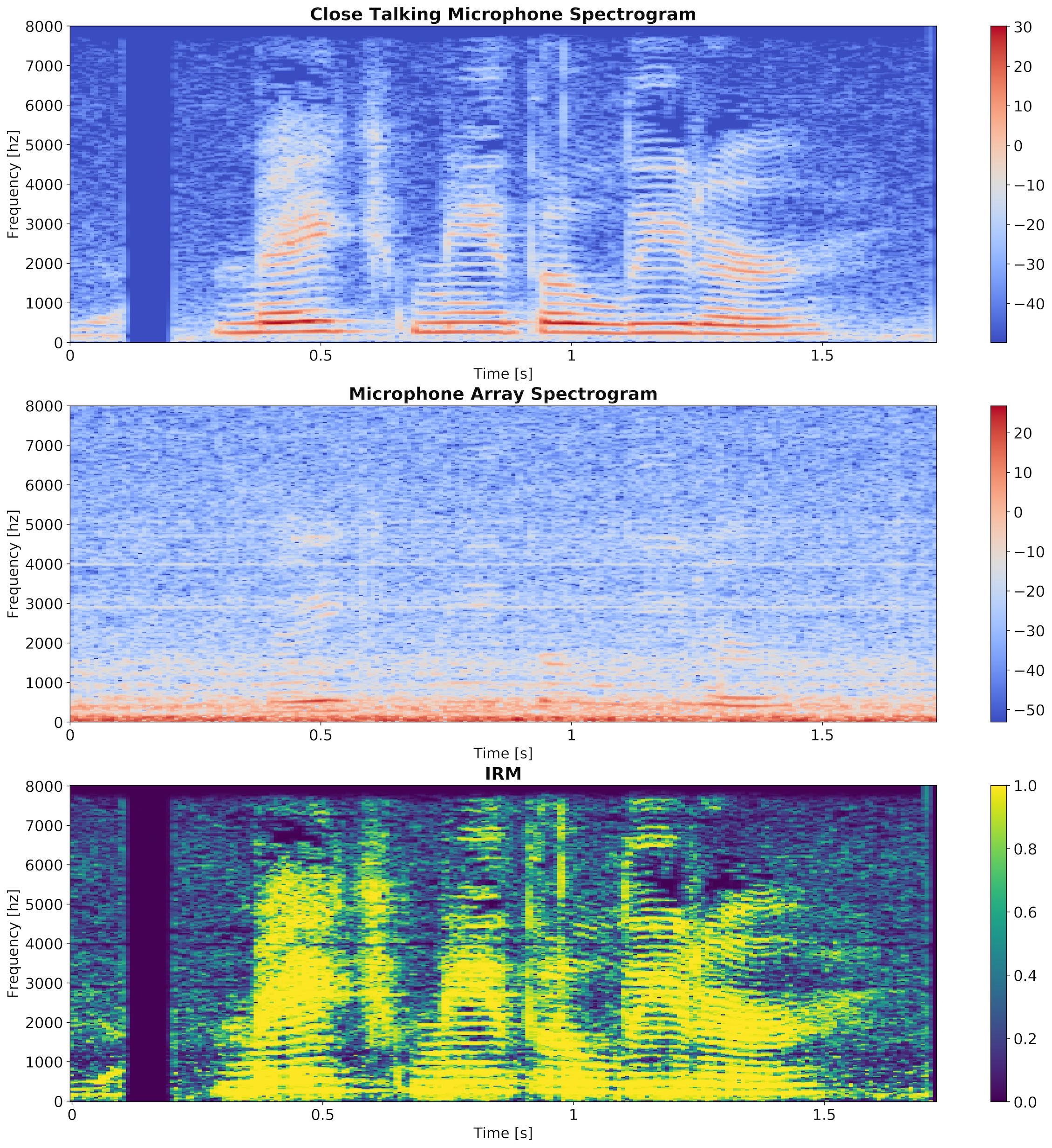}
      \caption{Plots of close talking microphone recording {\bf (top)}, noisy microphone array recording {\bf (middle)}, and the calculated ideal ratio mask (IRM) {\bf (bottom)}. The spectrogram of the close talking microphone is taken directly from the command recorded by the microphone attached to the Argo, while the spectrogram of the microphone array recording is taken after the command was broadcasted by the loud speaker}
      \label{fig::staticplots}
      \end{center}
\end{figure}

Figure \ref{fig::staticplots} shows the spectrograms of the close talking microphone and microphone array recordings, as well as the corresponding IRM produced by the pipeline. In the microphone array spectrogram, the speech is mostly corrupted by background noise. The proposed solution produces an IRM which discards the majority of noise present in the signal and reveals numerous speech features in the signal. The mask calculation can be performed in just under 2 seconds (as measured using C++ built in high resolution clock) on a Dell Precision with 16 cores, which suggests that the pipeline is viable in real world time constrained environments. 

Table \ref{table::staticerrors} presents the model average error and accuracy from the 8 distances tested for predictions with and without IRM. For each angle, the best 5 results of the 6 trials were used to exclude results affected by technical issues such as corrupted audio and software driver malfunctions, which occasionally compromised recording quality.

\begin{table}[h]
\caption{Average Error (\degree) and Accuracy (\%) within $5\degree$\\ with and without mask as a function of distance (m)} 
\begin{center}
\begin{tabular}{ccccc}
\toprule
\multirow{2}{*}{\textbf{Dist. (m)}} & \multicolumn{2}{c}{\textbf{No IRM}} & \multicolumn{2}{c}{\textbf{IRM}} \\
 & Avg. Error (\degree) & Acc. (\%) & Avg. Error (\degree)  & Acc. (\%)\\
\midrule
3 & 81.588 & 18.75 & 0.950 & 95.00\\
4 & 50.662 & 47.50 & 0.838 & 100.0\\
5 & 56.038 & 45.00 & 0.688 & 100.0\\
6 & 68.112 & 16.25 & 0.838 & 100.0\\
7 & 48.288 & 50.00 & 3.225 & 96.25\\
8 & 47.712 & 38.75 & 0.450 & 100.0\\
9 & 26.475 & 76.25 & 1.475 & 98.75\\
10 & 59.025 & 36.25 & 2.050 & 98.75\\
\bottomrule
\end{tabular}
\end{center}
\label{table::staticerrors}
\end{table}

\begin{figure}[thpb]
      \begin{center}
      \vspace{6pt}
      \includegraphics[width=\linewidth]{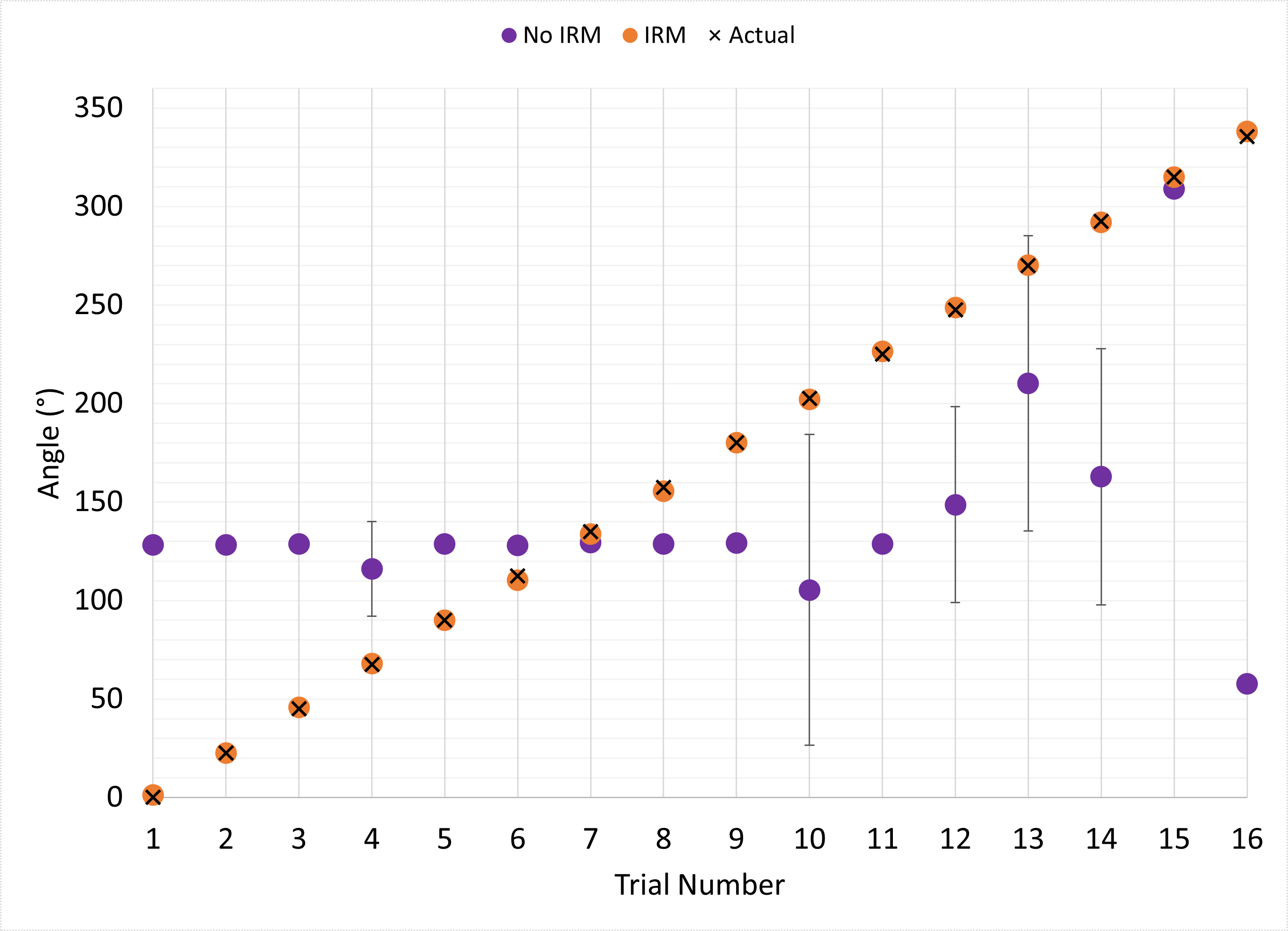}
      \caption{Model predictions with IRM, no IRM, and actual angles for trials done at 6m. The cluster of points at around 125 degrees is likely due to an air conditioning vent running in that general direction during trials.}
      \label{fig::errorplot}
      \end{center}
\end{figure}

Figure \ref{fig::errorplot} shows the pipeline predictions for trials done at $6$m. The predictions with the IRM and without IRM are shown in orange and purple respectively, and the true direction of the speaker is labeled with an cross symbol ($\times$). The results show the predictions of the pipeline with the IRM closely following the true direction of the speaker, while the pipeline predictions without the IRM cluster between $112\degree$ and $135\degree$. We believe the cluster of points in this region is because of an air conditioning vent running nearby during data collection in that direction. The predictions with the IRM also exhibit much lower standard deviations compared to no IRM predictions, which suggests the IRM is robust to potential sources of background noise. Overall, these findings show that the IRM is filtering a significant amount of noise which would otherwise prove detrimental to localization. 

The results presented in Table \ref{table::staticerrors} and Figure \ref{fig::errorplot} are derived only considering the farfield distance estimate during SRP-PHAT. The increase in accuracy with the inclusion of the nearfield was insignificant ($< 0.5\%$) and introduced just over a second of additional computational overhead. As a result, the decision was made to ignore the nearfield to allow the model to be more suited for real life scenarios. 

\subsection{Interference}

\begin{figure}[thpb]
      \begin{center}
      \vspace{6pt}
      \includegraphics[width=\linewidth]{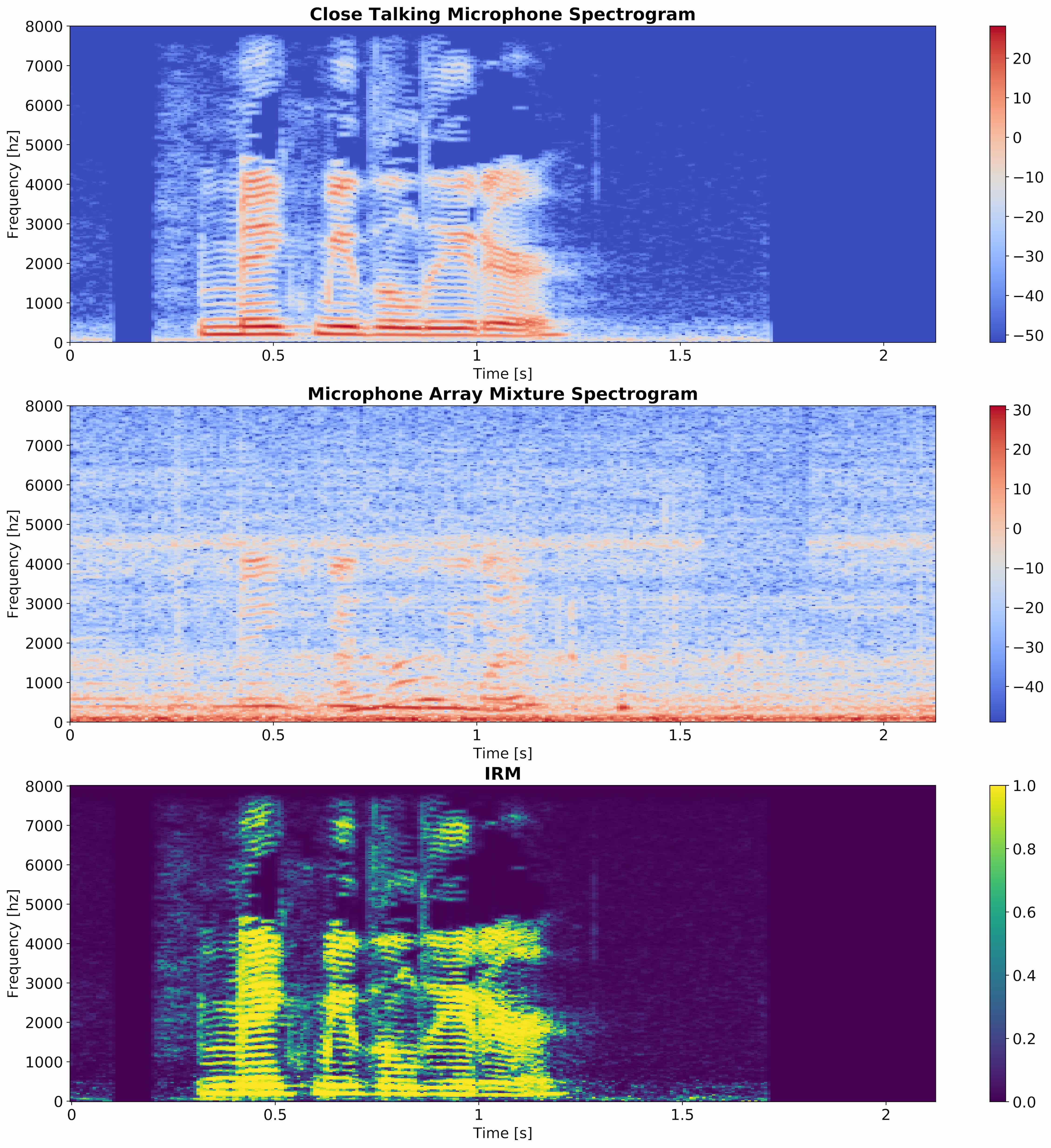}
      \caption{Plot of close talking microphone recording {\bf (top)}, noisy micro-
phone array recording {\bf (middle)}, and the calculated ideal ratio mask (IRM)
{\bf (bottom)}. The spectrogram of the close talking microphone is taken from the command recorded by the microphone attached to the Argo; however, the spectrogram of the microphone array mixture is plotted after mixing a mic array recording with another random voice command and white noise.}
      \label{fig::interferenceplot}
      \end{center}
\end{figure}

Figure \ref{fig::interferenceplot} shows the spectrograms of the close talking microphone and microphone array mixture recording, as well as the corresponding IRM produced by the pipeline. Despite the added noise from the non-target speech and white noise, the IRM is able to isolate the target speech and create a much clearer distinction between speech and background noise. 

Table \ref{table::interferenceerrors} summarizes the results of the pipeline for $3$ SNR values using no IRM, IRM ($\zeta[l,f]$ from (\ref{eq:zeta})), and the modified IRM ($\hat\zeta[l,f]$ from (\ref{eq:zetahat})), which will be denoted as IRM* moving forward. Results in table \ref{table::interferenceerrors} show that IRM* performs at least as good as IRM in high SNR environments, and outperforms the regular IRM at SNR values of $1.0$ and $10.0$. These findings suggest that IRM* is creating a clearer distinction between speech and noise, and increases pipeline robustness in both noise relaxed and dominant environments. We believe IRM*'s added robustness to inputs of low SNR is because even if the average SNR value of the input is low, some frequency bins still have locally high SNR and would behave in a similar manner as outlined before. While IRM* does add additional overhead required in the mask calculation, we believe it is valuable due to its improvement in accuracy and pipeline robustness.

\begin{table}[h]
\caption{Average Error (\degree) and Accuracy within 5\degree 
\\ with and without mask as a function of SNR (dB)} 
\begin{center}
\begin{tabular}{clcc}
\toprule
\textbf{SNR (dB)} & \textbf{Mask} & \textbf{Avg. Error (\degree)} & \textbf{Accuracy (\%)} \\
\midrule
\multirow{3}{*}{1.0} & None & 81.38 & 17.95 \\ 
& IRM & 10.72 & 89.74 \\ 
& IRM* & 4.00 & 94.87 \\
\midrule
\multirow{3}{*}{5.0} & None & 52.54 & 46.15 \\ 
& IRM & 3.15 & 97.43 \\ 
& IRM* & 2.87 & 97.43 \\
\midrule
\multirow{3}{*}{10.0} & None & 32.36 & 66.67 \\ 
& IRM & 14.54 & 89.74 \\ 
& IRM* & 7.08 & 94.88 \\
\bottomrule
\end{tabular}
\end{center}
\label{table::interferenceerrors}
\end{table}

\section{CONCLUSIONS}

This paper introduces an efficient pipeline to determine the direction of an operator in an unknown environment, using only a microphone array and close talking microphone. The framework relies on the calculation of an ideal ratio mask, a matrix that captures the signal present in the input that requires minimal computation. For robustness, a normalization term can be added to the mask calculation at a small cost to increase model stability in noise relaxed environments. Results obtained with a UGV and 16 microphone MA show more than a $50\%$ improvement in model average accuracy when the IRM is used compared to when it is not. Moreover, modifying the mask calculation can lead to an improvement of model average accuracy by approximately $5\%$ in both noise relaxed and dominant environments. This new pipeline brings opportunities to integrate object detection and leader follower routines on robots that require complex interaction with a speaker in an unknown environment. 

Future work will focus on integration of the system with person detection and gesture systems to allow for multi-sensor commands for robot control as well as development of an LLM-based voice control system allowing for rich context-aware commands to be issued to the UGV. We intend to expand the system to allow the UGV to be controlled by a small team of individuals to explore human-robot interaction effectiveness in defense scenarios.

\bibliography{references}
\bibliographystyle{ieeetr}




\end{document}